\documentclass[preprint,12pt]{elsarticle}

\usepackage{amssymb}
\usepackage{subfigure}
\usepackage{url}
\usepackage{color}
\usepackage{multirow}

\begin{document}

\begin{frontmatter}



\title{Hidden behind the obvious: misleading keywords and implicitly abusive language on social media}


\author[inst1]{Wenjie Yin}

\affiliation[inst1]{organization={School of Electronic Engineering and Computer Science, Queen Mary University of London},%Department and Organization
            addressline={Mile End Road}, 
            city={London},
            postcode={E1 4NS}, 
            %state={State One},
            country={United Kingdom}}

\author[inst1]{Arkaitz Zubiaga}

\begin{abstract}
 While social media offers freedom of self-expression, abusive language carry significant negative social impact. Driven by the importance of the issue, research in the automated detection of abusive language has witnessed growth and improvement. However, these detection models display a reliance on strongly indicative keywords, such as slurs and profanity. This means that they can falsely (1a) miss abuse without such keywords or (1b) flag non-abuse with such keywords, and that (2) they perform poorly on unseen data. Despite the recognition of these problems, gaps and inconsistencies remain in the literature. In this study, we analyse the impact of keywords from dataset construction to model behaviour in detail, with a focus on how models make mistakes on (1a) and (1b), and how (1a) and (1b) interact with (2). Through the analysis, we provide suggestions for future research to address all three problems.
\end{abstract}


% \begin{highlights}
% \end{highlights}

\begin{keyword}
 hate speech \sep abusive language \sep social media \sep text classification
\end{keyword}

\end{frontmatter}

\section{Introduction}

While social media provides a platform for all users to freely express themselves, cases of \textbf{offensive language} are not rare and can severely impact user experience and even the civility of a community \citep{nobata_abusive_2016}. When such offence is intentional or targeted, it is further considered \textbf{abuse} \citep{caselli_i_2020}. %> define, or move around from bottom
\textbf{Hate speech}, which is speech that directly attacks or promotes hate towards a group or an individual member based on their actual or perceived aspects of identity, such as ethnicity, religion, and sexual orientation \citep{waseem_hateful_2016,davidson_automated_2017,founta_large_2018,sharma_degree_2018}, is a sub-category of abuse that is identity-oriented \citep{vidgen2021introducing} and particularly harmful for silencing marginalised groups.\footnote{For a more elaborate comparison between similar concepts, see \cite{fortuna_survey_2018}, \cite{poletto_resources_2020}, \cite{banko_unified_2020}}

The problems caused by the posting of abusive and offensive language by social media users has increased the need both for analysing the phenomenon \cite{chandrasekharan2017you,kursuncu2019modeling,mathew2020hate} and for developing automated means for moderation of such content \cite{schmidt_survey_2017}. While methods for automated detection of abusive language are improving, the detection models share shortcomings that limit their practicality, due to their reliance on prominent lexical features, i.e. indicative keywords such as slurs and profanity. The effect of this reliance is two-fold. On one hand, they struggle with implicit expressions without such features and non-abusive speech with such features; 
on the other, the models have limited generalisability, i.e. models trained on one abusive language dataset don't perform well on other, unseen datasets \cite{arango_hate_2020}, where prominent features can be different. The two problems go hand-in-hand.

In this paper, we focus particularly on how the presence of keywords categorised as profanity or slurs can impact the tendency of models to label content as abusive, offensive or normal. 
To study this, we assess how decisions made in the creation of these datasets, including data sampling and annotation, can have an impact on the keyword and class distributions in the datasets, which in turn impacts model behaviour, both with detection and generalisation across datasets.

\subsection{Problem statement and research questions}
Abusive language detection methods are often built upon lexicon-based sampling methods, causing them to struggle with indirect forms of expressions \citep{han-tsvetkov-2020-fortifying}, and are easily misled by prominent words \citep{grondahl_all_2018}.

Challenges posed by keywords are two-fold: (1) non-abusive use of profanity and slurs, (2) implicit abuse without slurs or profanity.  Implicit abuse is the most commonly mentioned cause of false negatives in error analysis \citep{zhang_hate_2018,qian_leveraging_2018,basile_semeval-2019_2019,mozafari_bert-based_2020}. 
On the other hand, even state-of-the-art models, such as Perspective API,\footnote{\url{https://www.perspectiveapi.com/}} tend to flag non-abusive use of slurs \citep{badjatiya_stereotypical_2019, vidgen2021introducing}, which is common in certain language styles, such as African American English dialect \citep{sap_risk_2019}. 
These two challenges hurt the practicality of applying models as real-world moderation tools \citep{duarte_mixed_2018}: whilst cases of (1) may leave nonetheless-harmful content unaffected by moderation, cases of (2) may amplify the harm against minority groups instead of mitigating such harm as intended \citep{davidson_racial_2019}.

In addition to this, limited generalisability, demonstrated by the model performance drop when applied on unseen datasets, severely hurts the practical value of these automated detection models \citep{grondahl_all_2018,wiegand_detection_2019,swamy_studying_2019,arango_hate_2020,fortuna_how_2021, yin2021towards}. Models that suffer from this problem range from simpler classical machine learning models to recent state-of-the-art neural models \citep{badjatiya_stereotypical_2019, zhang_detecting_2018, bert/corr/abs-1810-04805}. What dataset factors contribute to better generalisability is thus an important and pressing issue.

Despite the recognition of the problems, there still exist considerable gaps in the understanding of them. On the relationship between keywords and model behaviour, most findings have been high-level, and the mechanism of the effect remains unclear. Dataset comparisons in generalisation studies mainly compared whole datasets and adopted a binary classification task, which oversimplify the problem and create inconsistent observations. Furthermore, the link between these two important and related issues is missing, such as the difference in, and factors that contribute to, generalisation on implicit and explicit abuse, as well as instances with misleading keywords that are not abusive.

\textbf{Research questions.} In this study, we aim to address the above-mentioned challenges brought by keywords (profanity and slurs), whilst distinguishing offensive and abusive (including hate speech). 

In particular, we ask: 

\begin{itemize}
    \item \textbf{RQ1:} How do approaches in dataset construction lead to different patterns of slurs and profanity presence in the dataset? 
    \item \textbf{RQ2:} How does such keyword presence in turn affect the detection model behaviour?
    \item \textbf{RQ3:} How does this effect differ when it comes to model generalisation?
\end{itemize}

To answer these questions, we make the following contributions:

\begin{itemize}
    \item We analyse the presence of keywords and the association between keywords and classes, comparing them across three commonly-studied datasets, and relating back to dataset construction in the analysis. 
    \item We then investigate the effect of such keyword presence on model behaviour, using a well-studied and strong model as a representative example.
    \item We perform in- and cross-dataset experiments to compare the effects on detection and generalisation.
\end{itemize}

As well as for dataset creation (involving collection, sampling and annotation) and for building automated detection models, our findings have important implications to enable large-scale analyses of behavioural and linguistic patterns linked to abusive and offensive language that incorporate nuanced abusive or non-abusive examples where the presence --or lack thereof-- of keywords entails the opposite of the expected meaning.

\section{Background} 

\subsection{Implicit expressions and non-abusive use of slurs and profanity}

Implicit expressions of abusive language and non-abusive use of slurs and profanity are the two sides of the same coin. As abusive language is associated with slurs and profanity, both its absence and presence can be misleading to the detection model -- when the speech is abusive and not so respectively. 

Both challenges have been recognised in abusive language
detection research, but usually separately and rather on a high level.

On one side, implicit abuse can be expressed through stereotypes, sarcasm, irony, humour, and metaphor \citep{sap_social_2019, mishra_tackling_2019, vidgen_challenges_2019}. It has been proposed that abusive language should be systematically classified into explicit and implicit \citep{waseem_understanding_2017}. Several subsequent studies have identified nuanced, implicit expression as a particularly important challenge in abusive language detection for future research to address \citep{aken_challenges_2018,duarte_mixed_2018, swamy_studying_2019}. Addressing implicit abuse is especially necessary for model explainability \citep{mishra_tackling_2019}. Manifested in the model behaviour, such implicit expressions are the most commonly mentioned cause of false negatives \citep{macavaney_hate_2019, mozafari_bert-based_2020, fortuna_how_2021}. 

The definition of explicitness and implicitness in the context of abusive language detection has been usually based on whether keywords -- slurs or profanity -- are present \citep{waseem_understanding_2017, caselli_i_2020, wiegand-etal-2021-implicitly-abusive}, although this definition is not equal to their linguistic or social definitions, as implicitness and explicitness are highly subjective notions \citep{vidgen_detecting_2020}.

The other side of the same coin -- non-abusive keyword use -- is equally important for abusive language detection. Profanity can be used for stylistic purposes or emphasis \citep{malmasi_detecting_2017}; some slurs have been reclaimed by targeted groups \citep{vidgen2021introducing}, and is common in African American English (AAE) dialect \citep{sap_risk_2019}. A model that falsely flags these instances as abuse could discriminate against minority groups that the model is intended to protect \citep{davidson_racial_2019}. Indeed, non-abusive slur and profanity use is a common cause of false positives \citep{aken_challenges_2018}; simply by adding the f-word can make positive statements carry a high ``toxicity'' score \citep{grondahl_all_2018}. 

Despite its importance in the abusive language detection task, it was only integrated into the taxonomy of abuse-surrounding phenomena recently \citep{vidgen2021introducing}.

Annotated implicit abuse and non-hateful slur use are very limited \citep{caselli_i_2020,vidgen2021introducing}.
Motivated by the lack of suitable data, recent studies have attempted to surface implicit abuse in unlabelled data, using initial samples with keywords \citep{gao_recognizing_2018} or through identifying ``influencial'' training instances  \citep{han-tsvetkov-2020-fortifying}. By providing more training instances with implicit abuse, both studies' approaches benefited model performance.

However, when it comes to the actual \textit{effect} of keywords, findings have been limited and largely high-level. In the error analysis of classification studies, it is generally considered that using keyword search and biased sampling contribute to a dataset containing more explicit expressions \citep{swamy_studying_2019, wiegand_detection_2019, caselli_i_2020, fortuna_how_2021}, which can lead to a higher recall of the positive class \citep{graumas_twitter-based_2019} but overall lower F1 \citep{razo_investigating_2020} in binary classification. Many questions still remain: how annotation plays a role in the process, how the resulting keyword presence in the data impact the detection of implicit and explicit expressions differently, how implicitness interact with other factors, etc.

\subsection{Generalisability in abusive language detection}

Generalisation refers to how well a machine learning model performs on previously unseen data, which assesses its ability of capturing the real relationship between features and expected outputs \citep{Goodfellow-et-al-2016}. 

Recently, the generalisability of abusive language detection models have received increasing attention. Cross-dataset testing -- evaluating models on a different dataset than the one(s) they were trained on -- has revealed that model performance is severely over-estimated when evaluated only on the set-aside ``test set'' of the same dataset. A recent study has summarised the impact of model and dataset factors on cross-dataset performance \citep{yin2021towards}. Across different cross-dataset studies, the macro-averaged F1 scores most commonly drop by between 20 and 40 points when the model is applied on a different dataset \citep{grondahl_all_2018,wiegand_detection_2019,swamy_studying_2019,arango_hate_2020,fortuna_how_2021}.  
Models that suffer from a significant performance drop when applied cross-dataset range from classical machine learning models to the state-of-the-art neural models \citep{badjatiya_stereotypical_2019, zhang_detecting_2018, bert/corr/abs-1810-04805}.

Nonetheless, fine-tuning large language models that have been pre-trained extensively, with BERT \citep{bert/corr/abs-1810-04805} as a representative example, seems to be relatively more generalisable \citep{pamungkas_misogyny_2020, fortuna_how_2021}. 

Studies on factors that affect generalisation have mostly compared entire datasets. Similar datasets generalise better to each other, with the similarity attributed to search terms for sampling \citep{swamy_studying_2019}, topics \citep{nejadgholi_cross-dataset_2020, pamungkas_misogyny_2020}, and class label definitions \citep{pamungkas_misogyny_2020, fortuna_how_2021}. When it comes to what produces high-quality generalisable data in general, wider coverage of abuse phenomena, including topics, is believed to be beneficial \citep{pamungkas_cross-domain_2019, nejadgholi_cross-dataset_2020}. A more broadly defined positive class is also perceived as more generalisable to a narrowly defined one than the other way round in binary classification \citep{nejadgholi_cross-dataset_2020, fortuna_how_2021}. Surprisingly, the effect of the data size is rather limited, compared to other factors \citep{nejadgholi_cross-dataset_2020, fortuna_how_2021}.
Inconsistency exists in the literature, when it comes to the effect of class proportions: while some observed that a larger proportion of the positive class makes a dataset more generalisable \citep{karan_cross-domain_2018, swamy_studying_2019}, others found the opposite \citep{wiegand_detection_2019} or could not confirm its effect \citep{fortuna_how_2021}. One explanation is that the balance between true positive and true negative is not reflected in the overall performance \citep{nejadgholi_cross-dataset_2020}. 

Findings on the relationship between the presence of keywords and generalisation are limited to a few isolated observations. Some hold that containing more explicit expressions makes a dataset more generalisable \cite{wiegand_detection_2019}; others observed that most non-offensive instances with keywords in \cite{davidson_automated_2017} were mislabelled as offensive by a model trained on \cite{zampieri_predicting_2019}, and attributed this to the high frequency of keywords in the former \citep{swamy_studying_2019}. Thus, to fill in the gaps, systematic investigations on this topic is needed.

\subsection{Offensive language vs. abuse}\label{ssec:offensive_abuse}

There exists a key difference between offensive and abusive language: abusive language has a strong component of intentionality; the definition of offensiveness has more emphasis on lexical content and the receiver's emotional response. Hate speech, with a strong intention to ``direct attack or promote hate'', thus falls under abusive language. 
Experts can distinguish abusive and offensive, both conceptually and in practice during annotation \citep{caselli_i_2020}.
However, both are used as umbrella terms for harmful content in the context of automatic detection studies, and these two terms are often confused, especially by crowd annotators. In a large-scale crowd-annotated dataset  
\citep{founta_large_2018}, the annotations for ``abusive'' and ``offensive'' were so similar that the two class labels were combined in the end. 

This distinction carries significant practical value. On offensive language, a purely lexicon-based detection model can achieve competitive performance \citep{pedersen2019duluth}, while abuse is captured by lexical features less \citep{caselli_i_2020}. Thus, distinguishing abuse and offensive language can reveal more insights into implicit expressions.

~\\
In summary, despite their importance, inconsistency and unanswered questions still largely remain, in both the challenges posed by keywords and model generalisability. Furthermore, existing studies on these two topics have always worked with binary classification without distinguishing offensive language and abuse, limiting the practical value.

Thus, our study addresses the gap in the literature by providing an all-round and in-depth analysis of the challenge posed by keywords -- unifying the two sides of the same coin and following the entire chain of effect: from sampling and annotation to the data, then finally to model behaviour.
By extending the above analysis from in-dataset detection to cross-dataset generalisation, we offer a new perspective of looking at model generalisation. 
We include but distinguish both abuse and offensive language, enabling insights into fine-grained model behaviour and better understanding of implicitness and keyword use. As a result, we clarify confusions in the interacting factors seen in previous studies.

\section{Materials}

\subsection{Definitions}

Table \ref{tab:definitions} summarises the main concepts used in this study. We consider three main types of nature of speech -- abuse, offensive, and normal, separating abuse and offensive with intentionality (\S \ref{ssec:offensive_abuse}) and including hate speech as a special case of identity-oriented abuse \citep{vidgen2021introducing}. 

Following the definitions of previous studies \citep{waseem_understanding_2017,caselli_i_2020, wiegand-etal-2021-implicitly-abusive}, whether an instance is implicit or explicit is then dependent on the presence of keywords, which can be any slur or profanity: if an instance of speech, whilst being abuse or offensive, contains more than one keyword, it is explicit. If keywords are present without the whole instance of speech being offensive or abusive, we consider it non-abusive keyword use, similar to \cite{vidgen2021introducing}. 

Our study focuses on all types of indirect expressions: implicit abuse, implicit offensive language, and non-abusive use of keywords.

\begin{table}[htp]
\small
\centering
\begin{tabular}{p{1.3cm}|p{2.5cm}p{4cm}p{4.3cm}}
\hline
                & Abuse (incl. Hate speech)         & Offensive                   & Normal                       \\ 
\hline
With keyword    & Explicit abuse & Explicit offensive language & Non-abusive use of keywords  \\
Without keyword & Implicit abuse & Implicit offensive language & /                            \\
\hline
\end{tabular}
\caption{Definition of implicit, explicit, non-abusive use in relation to the nature of speech. }
\label{tab:definitions}
\end{table}

\subsection{Resources}

We make use of two types of resources for our research: (1) a set of abusive language datasets labelled as abuse, offensive or normal, and (2) collections of keywords that enable us to distinguish, within the datasets, the cases that make explicit use of these keywords. By using the keywords to find matches, we break down the datasets into two subsets: ``\textbf{Any keyword}'' and ``\textbf{No keyword}''. When the ``\textbf{No keyword}'' subset overlaps with either the abuse or offensive label, we deem these \textbf{implicit abuse} and \textbf{implicit offensive language}, respectively.

\subsubsection{Datasets}
% > 

We chose to use three multi-class datasets: \textit{AbuseEval} \citep{zampieri_semeval-2019_2019, caselli_i_2020}, \textit{Founta} \citep{founta_large_2018} and \textit{Founta} \citep{founta_large_2018}. As opposed to the vast majority of existing datasets providing binary labels (abuse vs not), these three datasets were selected for enabling distinction of the three categories of our interest:  (1) abuse, which subsumes hate speech, (2) offensive, and (3) normal. 

The original class labels needed adapting slightly to enable comparative analysis across datasets by mapping them into the above three classes:

\begin{itemize} % > exact words
\item \textit{AbuseEval}. ``Abuse'' were used as-is. Instances that fall under ``offensive'' but not ``abuse'' were used as ``offensive''. 
\item \textit{Davidson}. Original classes (``hate'', ``offensive (but not hate)'', ``neither'') were directly mapped into (``abuse'', ``offensive'', ``normal''). 
\item \textit{Founta}. ``Spam'' and ``normal'' were combined into ``normal''. We made the decision based on the other two datasets -- both \textit{Davidson} and \textit{AbuseEval}'s ``normal'' classes contain instances that would be considered ``spam'' in \textit{Founta}\footnote{Such as ``I added a video to a USER playlist URL ...'', `` Charlie Sheen engaged to porn star URL ... ''}. ``Hate'' was mapped into ``abuse''. ``abusive'' was renamed to ``offensive''. We made this decision because their definition of ``abusive'' does not mention any intentionality and is hardly indistinguishable from that of ``offensive''; the annotators could not distinguish them, either.
\end{itemize}

The class labels after mapping are distributed as shown in Table \ref{tab:datasets}.

\begin{table}
\small
\centering
\begin{tabular}{l|ccc|l} 
\hline
            & Abuse & Offensive & Normal & Total \\ 
\hline
AbusEval    & 2927  & 1713      & 9460   & 14100 \\ 
\hline
Davidson    & 1430  & 19190     & 4163   & 24783 \\ 
\hline
Founta      & 4965  & 27150     & 67881  & 99996 \\
\hline
\end{tabular}
\caption{Statistics of the three datasets used in our study.}
\label{tab:datasets}
\end{table}

There are two main differences to notice in the datasets.
\textit{Founta} is a few times larger than the other two, and \textit{Davidson} close to twice the size of \textit{AbuseEval}. All three datasets are imbalanced, but in different ways: The majority class is Normal in \textit{Founta} and \textit{AbuseEval} as in most other abusive language datasets, but Offensive in \textit{Davidson}. The smallest class is Abuse for both \textit{Davidson} and \textit{Founta}, but Offensive for \textit{AbuseEval}.

\subsubsection{Keywords}
We gathered widely used sets of keywords that can be categorised as either slurs or profanity. We use the Hatebase\footnote{https://www.hatebase.org/} lexicon to cover 1532 slurs, and the No  Swearing\footnote{https://www.noswearing.com/dictionary/} lexicon to cover 298 profanity words, after excluding from the latter those that are also considered slurs. In the case of the 1532 Hatebase slurs, we also preserve information on what attribute of the victim (topic) the slur is targeting at, e.g. ethnicity or religion -- the latter enables an analysis by topic in \S \ref{ssec:topic}.

\section{The source of the problem: slurs and profanity in the data, and their association to offensiveness and abusiveness}

We first analyse the \textbf{Any keyword} and \textbf{No keyword} subsets to understand (1) how implicit abuse and offensive language are manifested in the datasets, and (2) to assess the presence of keywords in the normal class.

We show that certain datasets are (a) more \textbf{keyword-intensive} --containing keywords more often overall--, and certain datasets are (b) more \textbf{keyword-dependent} --the association between keyword presence and class labels is stronger. It is important to note that one does not necessarily determine the other. We then break the keyword presence down by possible topics -- about ethnicity, gender, ... or just general swearing. 

\subsection{Keyword presence and its association to abusiveness and offensiveness}\label{subsec:keyword_notopic}

We show in Table \ref{tab:presence} the breakdown of the three datasets by class label and keyword presence. 

Overall, regardless of the nature of speech, the overwhelming majority of \textit{Davidson} posts contain keywords, making it the most keyword-intensive, in contrast to less than half for the other two datasets.

\begin{table}
\small
\centering
\begin{tabular}{l|lll|l|c} 
\hline
         & Abuse & Offensive & Normal & Total  & $\chi^2$ \\ 
\hline
\multicolumn{6}{c}{\textbf{AbuseEval}} \\ 
\hline
Any keyword     & 764   & 707       & \textcolor[rgb]{0.16,0.57,0.16}{\textit{583}}    & 2054   & \multirow{2}{*}{1831.459} \\
No keyword  & \textcolor{red}{\underline{2163}}  & \textcolor{red}{\underline{1006}}      & 8877   & 12046  &\\ 
\hline
\multicolumn{6}{c}{\textbf{Davidson}} \\ 
\hline
Any keyword     & 1408  & 19163     & \textcolor[rgb]{0.16,0.57,0.16}{\textit{3992}}   & 24563  & \multirow{2}{*}{619.159} \\
No keyword  & \textcolor{red}{\underline{22}}    & \textcolor{red}{\underline{27}}        & 171    & 220    &\\ 
\hline
\multicolumn{6}{c}{\textbf{Founta}} \\ 
\hline
Any keyword     & 2673  & 23755     & \textcolor[rgb]{0.16,0.57,0.16}{\textit{5517}}   & 31945  & \multirow{2}{*}{57343.064} \\
No keyword  & \textcolor{red}{\underline{2292}}  & \textcolor{red}{\underline{3395}}      & 62364  & 68051  &\\ 
\hline
\end{tabular}
\caption{The three datasets broken down by class labels and whether having keywords. The chi-squared statistics shows the dependency between keyword presence and class labels. All p-values < 0.001. }%*: Where abuse is the special case of hate speech. }
\label{tab:presence}
\end{table}

Having at least one keyword means that an instance is far more likely to be offensive or abusive than innocent. A chi-squared ($\chi^2$) test of dependence further confirms this. 
A dataset being more keyword-intensive overall does not mean that the class labels in that dataset are more keyword-dependent, comparing the ratio of instances with keywords and the chi-squared values.

The proportion of implicit abuse and implicit offensive language (red, underlined), and non-abusive use of keywords (green, italic) in a dataset depends on both overall keyword-intensity and class keyword-dependency.

\subsection{A closer look into the topics}
\label{ssec:topic}

\begin{figure}[htp]
    \centering
    \subfigure[AbuseEval]{\includegraphics[scale=0.17]{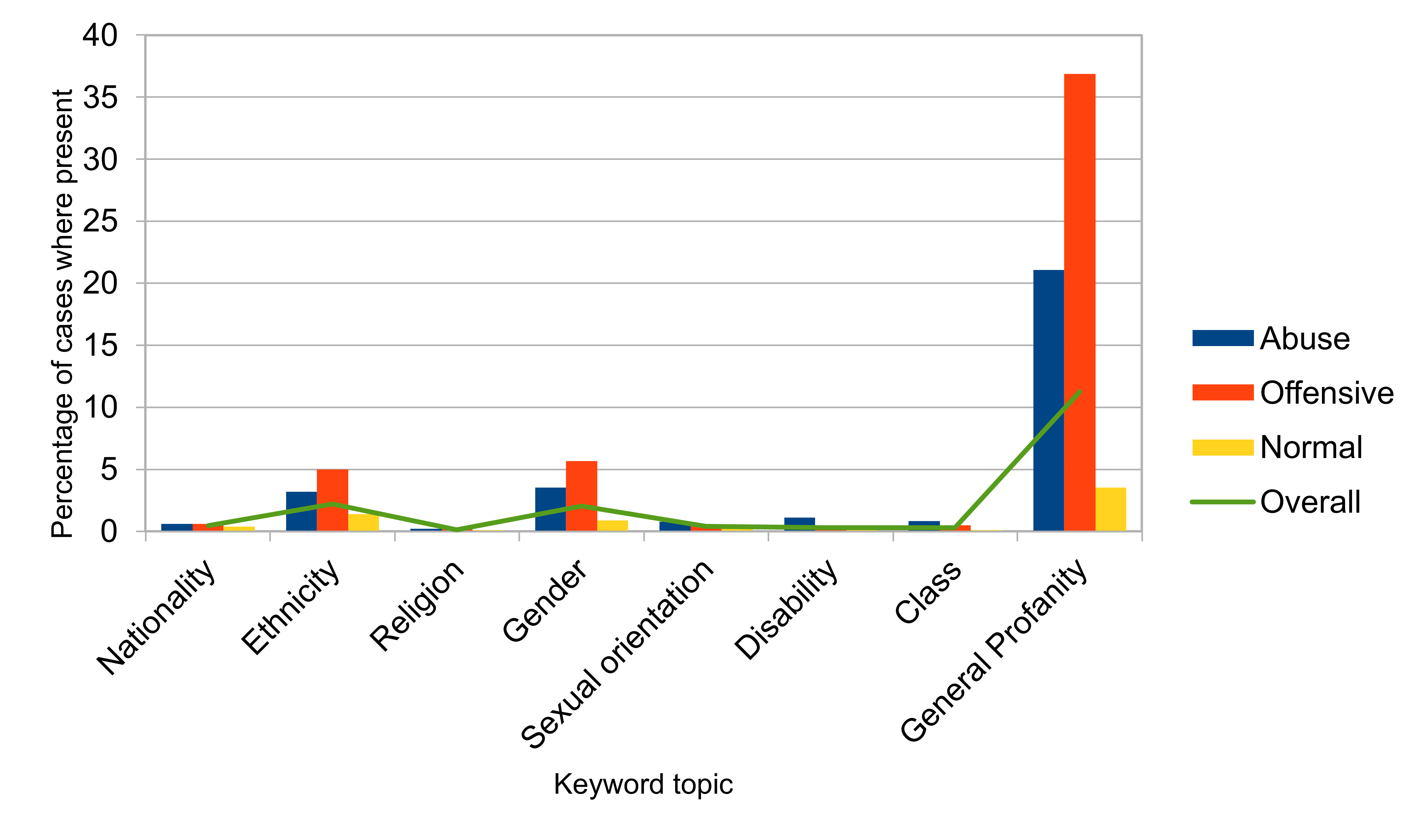}
    \label{subfig:topics_abuseeval}}
    \subfigure[Davidson]{\includegraphics[scale=0.17]{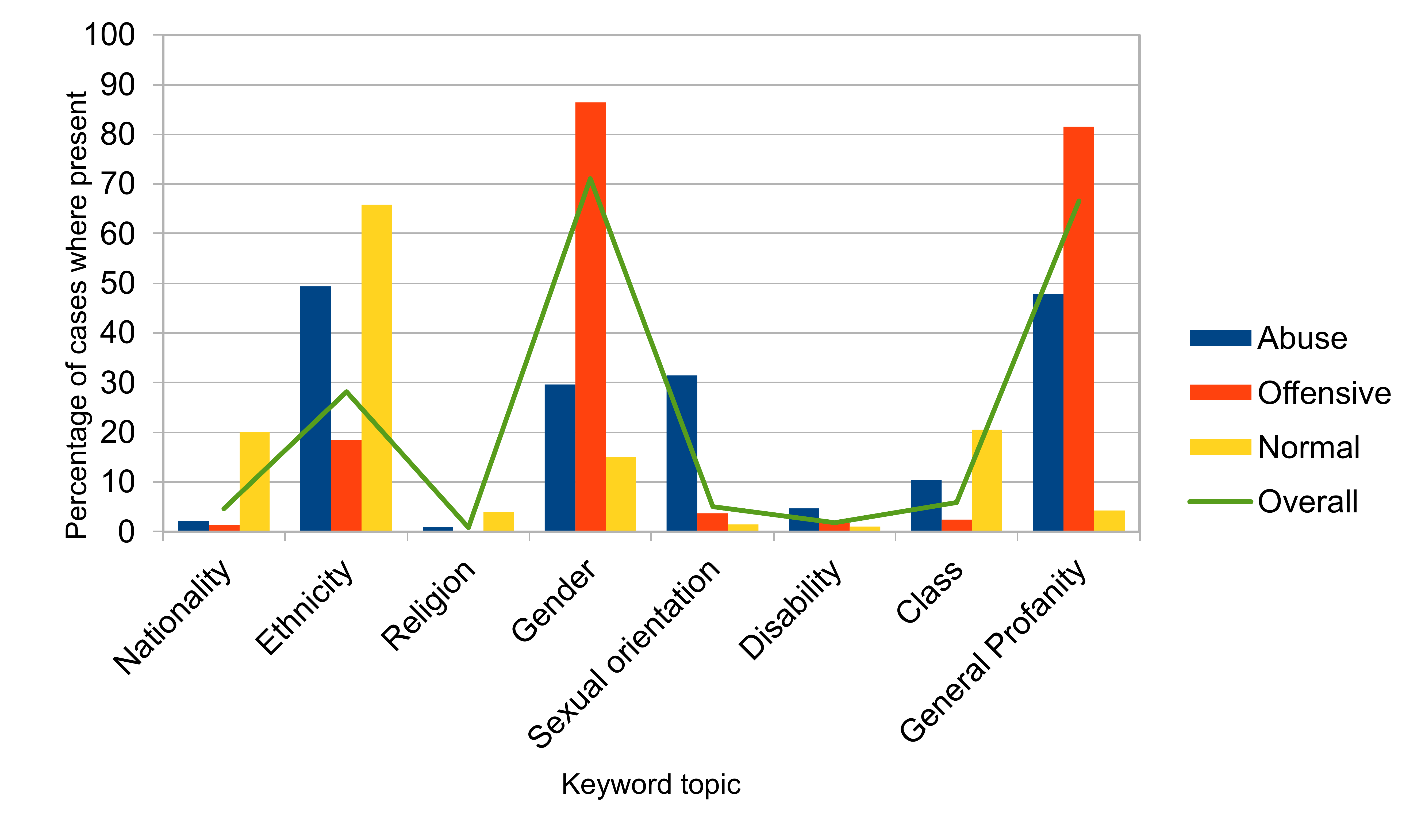}
    \label{subfig:topics_davidson}}
    \subfigure[Founta]{\includegraphics[scale=0.17]{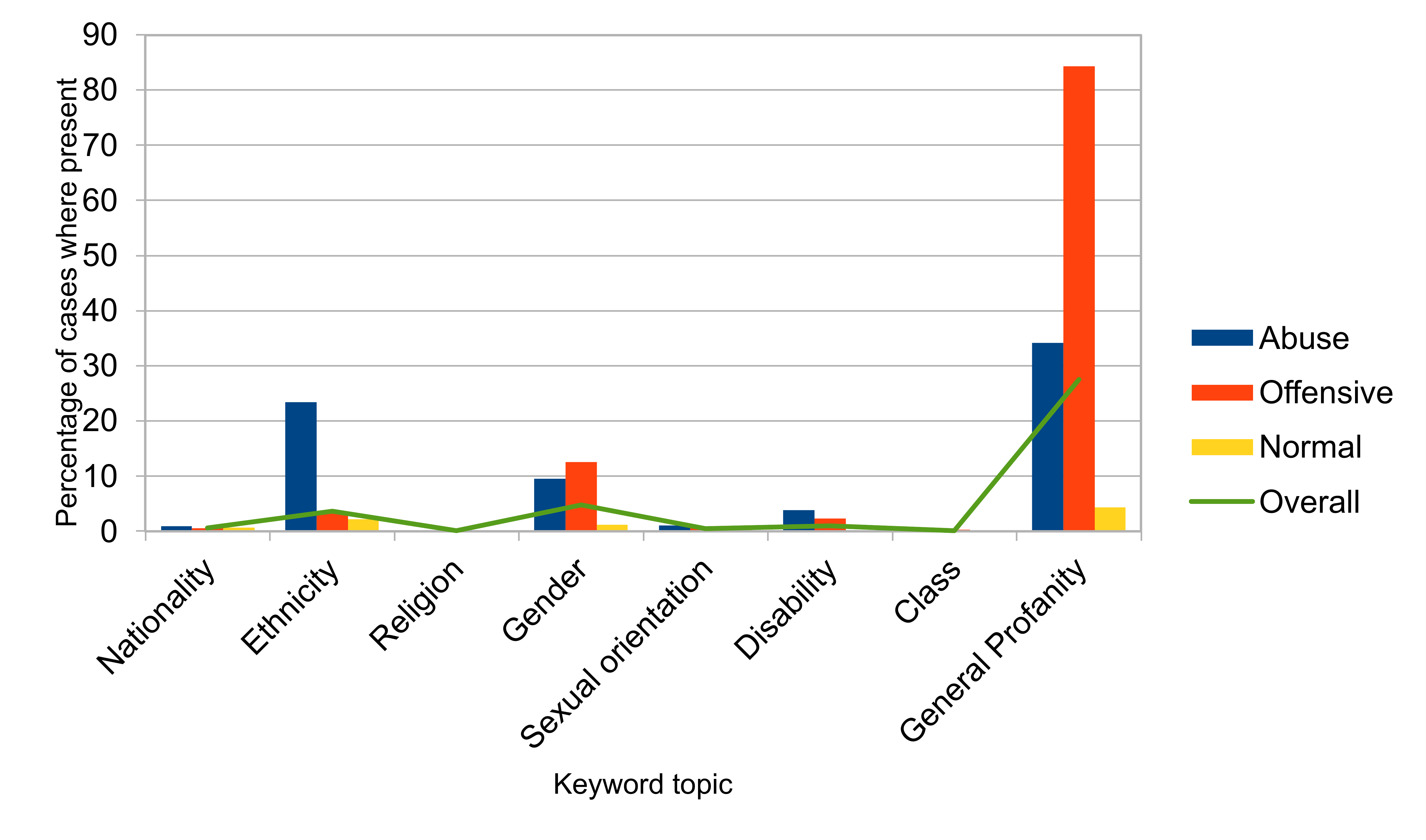}
    \label{subfig:topics_founta}}    
    \caption{Keyword topics in the three datasets, shown as the percentage of instances in the whole dataset or under each class that contain any keyword of that topic. Note that the y-axis is different to prioritise easier comparison across topics in each dataset. }
    \label{fig:topics}
\end{figure}

The topics of the keywords present -- the overall presence in the whole dataset and the relative presence across different types of speech -- are shown in Figure \ref{fig:topics}. 

Regardless of the nature of speech and the specific dataset, general profanity is the most common type of keyword, followed by slurs related to ethnicity and gender. At the same time, \textit{Davidson} stands out from the other two datasets  -- general profanity appears much more frequently than any slurs in both \textit{Founta} and \textit{AbuseEval}, while, in \textit{Davidson}, gender- and ethnicity-related slurs are also very common, with the former even more so than general profanity. For any keyword topic, the frequency is the highest in \textit{Davidson}, the most keyword-intensive dataset.

Certain topics are associated with certain class labels. One might expect ethnicity-related slurs to be associated with abuse and especially hate speech in the datasets, as they have been with racism historically. Interestingly, only \textit{Founta} displayed such tendency unequivocally: under the ``abuse'' class, which is formed by annotated hate speech, ethnicity-related slurs are about five times more likely to appear than the other two classes. In clear contrast, such slurs are also commonly found under the ``normal'' class in \textit{Davidson}, meaning they are used in innocent settings. Nonetheless, they are still much more common in abusive than offensive language. The difference is much smaller in \textit{AbuseEval}, with these slurs taking up only a small fraction in all classes. 
The pattern across classes for gender-related slurs is much more consistent across the three datasets. They seldom appear in the normal cases; they are more common in offensive language that is not considered abuse or hate speech. General profanity follows a similar trend, except the frequency is much higher across all classes.
Some types of slurs cover a noticeable proportion only in \textit{Davidson}: Sexual orientation-related slurs are fairly common in hate speech, but very rare in non-hate offensive language and normal speech; nationality- and class- related ones are most common in normal speech. 

All in all, the different patterns across topics show how different slurs and profanity are used and perceived and echo with the overall keyword presence and its association with offensive and abusive speech (\S \ref{subsec:keyword_notopic}).

\subsection{Discussions: from dataset construction to model training} \label{ssec:discussion1}

Overall presence of keywords and the association between keywords and class labels can be traced back to how the datasets were built, and will have effects on any classification model trained on the datasets. In what follows we link our analysis with the sampling and annotation strategies, as well as discuss the expected impact on abusive language detection models.

\subsubsection{Sampling}

Sampling -- how the initial data is gathered before annotation -- affects how keyword-intensive the dataset is and the distribution of the topics, and reflects the domain of the dataset by showing how the data construction is motivated.

Offensive language generally represent less than 3\% of social media content \citep{zampieri_semeval-2019_2019,founta_large_2018}. Thus, all datasets apply some process, such as text search, to increase the proportion of offensive and abusive content.

There are two key components which contribute to how biased towards keywords the sampling process is: the approach and the criteria. Sampling approaches in offensive language datasets can be divided into boosted random sampling and biased sampling \citep{wiegand_detection_2019}. While the former approach applies the criteria after drawing an initial random sample, the latter draws a biased sample with the criteria. The criteria can be compared in terms of whether inherently offensive terms are used for search.

Biased sampling was applied on \textit{Davidson}. An initial sample was drawn through lexicon search with Hatebase, and timelines from users identified in the lexicon search were also included. This very focused approach resulted in the dataset having the most intensity and broad coverage of slurs and profanity overall.
On \textit{Founta}, Hatebase was similarly used for lexicon search, in addition to No Swearing. Negative sentiment was also part of the criteria. Despite the direct search of slurs and profanity, a boosted random sampling approach was taken. As a result, it contains such keywords much less frequently than \textit{Davidson}, which in turn makes the dataset a lot less keyword-intensive, with a lot of clean cases without any keyword or offensive language.
\textit{AbuseEval} is a re-annotated version of the \textit{OLID} dataset \citep{zampieri_predicting_2019}, and therefore inherits the sampling criteria of the latter. The text search criteria applied in this case is a lot less direct, such as ``you are'', ``she is'', ``MAGA'', ``gun control''. They also looked for replies to extreme-right news and posts that get filtered out by safe search. These less direct criteria reflect that the dataset aims to cover a broader spectrum of and less extreme types of offensive language. The presence of slurs and profanity keywords in the resulting dataset is thus much less frequent than the other two hate speech datasets, making it the least keyword-intensive among the three. Nonetheless, interestingly, the topic distribution is very similar to that of \textit{Founta} (Figure \ref{fig:topics}).

What's more, sampling through text search has an unintended effect on class distributions. It fulfils the intended purpose to boost the proportion of offensive or abusive posts, indeed. As shown in Table \ref{tab:presence}, all three datasets contain much more offensive or abusive posts than social media generally, although using boosted random sampling instead of biased sampling (\textit{Founta}) or using terms not inherently offensive for biased sampling (\textit{AbuseEval}) makes the effect milder.  On the other hand, sampling through text search reduces the actual proportion of actually abusive posts within the offensive ones. This is because abuse is less likely to contain keywords, as shown in all three datasets, and as pointed out by \citep{caselli_i_2020} that abuse depends less on lexical features. Comparing the three datasets, the more keyword-intensive a dataset is, the smaller the ``abuse'' class is compared to ``offensive'', with \textit{Davidson} being the most so and \textit{AbuseEval} being the least. 

\subsubsection{Annotation}

General profanity can be used for stylistic purposes or emphasis \citep{malmasi_detecting_2017}; some slurs have been reclaimed by targeted groups \citep{vidgen2021introducing}, and is common in African American English (AAE) dialect \citep{sap_risk_2019}. It is thus necessary to distinguish keyword use and actual offensiveness or abusiveness.

Inconsistencies in definitions across datasets challenge model generalisability \citep{fortuna_toxic_2020}. Datasets also vary by how specific the guidelines are, such as a more detailed explanation of the definition and clarifications on edge cases.
Then, annotators differ by whether they were crowd-sourced or experts. Generally, expert-annotated data are considered of higher quality \citep{vidgen_directions_2020} and produce better-performing classification models \citep{waseem_are_2016}.

Among the three datasets, \textit{Founta} annotators were allowed the most freedom as the annotation instructions and taxonomy weren't fully-fledged from the beginning, but were progressively completed as annotations were collected. The annotations were completely crowd-sourced; annotators first carried out an exploratory round of annotations, based on very brief definitions of a range of possibly overlapping concepts, after which the final class labels were then defined. This resulted in annotations heavily based on keywords (Table \ref{tab:presence}). Thus, although also focusing on the specific abuse type of hate speech during annotation like \textit{Davidson}, the annotated ``hate speech'' label in the original dataset is expected to cover a wider range of phenomena, including some instances that contain slurs but are not hate speech. 
For example, in \textit{Founta}, ``This what happens when you separate yo self from ni\*\*as who don ' t eat they food cold. You FLOURISH...'' was labelled as Hateful, while the n-word was used in a reclaimed manner. In comparison, similar usage of the n-word in \textit{Davidson} such as ``I ain ' t never seen a bitch so obsessed with they nigga (emoji) I ' m obsessed with mine (emoji)'' was consistently labelled as Offensive.

The other two datasets had defined the class labels before the annotation process.
Although also completely crowd-annotated, the annotators of \textit{Davidson} received more intensive instructions: a paragraph explaining in detail along the definition of each class label; they were also explicitly asked to not base their judgements solely on the words in isolation. This specific instruction is expected to make its ``abuse'' class the most specific among the three datasets, focusing on hate speech and with annotators' bias towards keywords reduced. It is reflected in the data by having the least inter-dependency between keywords and class labels among the three datasets. 
\textit{AbuseEval} has both crowd and expert annotations: in two separate studies, experts annotated the abusiveness \citep{caselli_i_2020} of originally crowd-annotated offensive posts \citep{zampieri_predicting_2019}. 
The definition of offensiveness is loose and broad, covering ``any form
of non-acceptable language (profanity) or a
targeted offense''. Although carrying a broader definition of abuse without focusing on a subtype like the other two datasets, abusiveness was annotated with clearer instructions in the form of a decision-tree. This resulted in an overall moderately keyword-dependent dataset. 

On the other hand, common trends are found when it comes to the keyword topics and perceived hatefulness in the two hate speech datasets (\textit{Davidson}, \textit{Founta}). While instances labelled as offensive contain gender-related slurs much more often than abuse, those labelled abuse much more frequently contain ethnicity-related slurs. There are two possible explanations to this: Indeed, the authors \citep{davidson_automated_2017} noticed that annotators tended to perceive racism and homophobia as hate speech, but sexism as only offensive, consistent with the findings of an earlier dataset study \citep{waseem_are_2016}. % Cross-dataset experiments also showed that ``hate speech 
This likely reflects how the western societies, where the research is focused, perceive different types of abuse. For instance, European countries most commonly centre their legal definition of hate speech around race, followed by religion and sexual orientation\footnote{https://www.legal-project.org/issues/european-hate-speech-laws}.

To summarise, crowd annotators, who reflect the general public, display a tendency to rely on keywords, while experts rely on keywords less. Offensive and abusive are seen as fundamentally different by experts, but are confused by crowd annotators. Instructions for crowd workers to actively consider the context of words used, as with instruction to consider the dialect of the speaker \citep{sap_risk_2019}, reduce biases induced by slurs. Some topics are perceived more abusive/hateful than others.

\subsubsection{Implications on classification model training}\label{subsubsec:implication_training}

Based on the analysis above, we can expect the data to have effects on classification model training, including both utilising pre-trained models and fine-tuning to the task.

First of all, a model trained on an imbalanced dataset is expected to display a tendency to predict the majority class the best. 
Then, keywords' overall presence and association with the class labels would make classification difficult. A pre-trained model would already have encoded offensive meanings of slurs and profanity. Thus, keyword-intensive data, such as \textit{Davidson}, can mislead a pre-trained model even before exposing it to keyword-label combinations. During the fine-tuning stage, the association of keywords with offensive and abusive class labels would be further integrated into the model, with the effect being the strongest in the more keyword-dependent datasets, such as \textit{Founta}.

Furthermore, models would struggle to generalise to unseen datasets, although generalisation is expected to better in dataset pairs which are more similar. In terms of keywords, \textit{Founta} and \textit{AbuseEval} are more similar, both being more keyword-dependent, less keyword-intensive, and having similar keyword topic distributions. In terms of class labels, \textit{Davidson} and \textit{Founta} both focusing on a specific type of abuse, hate speech. Nonetheless, the guidelines of the latter are broader, making it relatively similar to \textit{AbuseEval}, where both ``abusive'' and ``offensive'' are umbrella terms. 
Considering both factors, it is thus expected that generalisation between \textit{Davidson} and \textit{AbuseEval} would be the most challenging.

\section{Challenges reflected in the classification model: when the words present don't match the underlying meaning}

In the last section, we analysed the presence and association of keywords in the datasets, related them back to dataset building, and hypothesised what impact they would have on model training. In this section, we assess such impact in detail through both in- and cross-dataset experiments. We also extend these discussions to a novel generalisation scenario where two sources of out-of-domain data are combined for training.

The model we use for this assessment, BERT \citep{bert/corr/abs-1810-04805}, is commonly used in abusive language detection and achieves strong results in shared tasks (with in-dataset evaluation) \citep{zampieri_semeval-2020_2020,fersini_ami_2020} and generalisation studies (with cross-dataset evaluation) \citep{pamungkas_misogyny_2020, fortuna_how_2021}. Thus, although we only consider one model design, the results are expected to reveal common issues in most if not all abusive language detection models.

The classification model we use is BERT-base-uncased with transformer layers and a subsequent pooling layer all initialised with the pre-trained weights obtained from Huggingface\footnote{http://huggingface.co/transformers}. After the pooling layer, a fully-connected layer, randomly initialised, maps the pooled representation to a class prediction through a Softmax function. 

We settled on a learning rate of $1e^{-5}$, maximum sequence length of 70 after hyperparameter experiments on the validation set, similar to the settings of a previous study \citep{swamy_studying_2019}. We saved model checkpoints every 1000 steps and performed early stopping after 4000 steps of no improvement over validation macro-F1 with a maximum budget of 20000 steps. The rest of hyperparameters were kept as default. 
Performance metrics reported are all means computed from the 8 models.

The mean macro-averaged F1 scores are shown in Table \ref{tab:macro_f1}. For the remainder of this section, we focus on performance metrics of specific class labels in relation to keyword presence.

\begin{table}
\centering
\begin{tabular}{l|lll} 
\hline
Training / Evaluation & AbuseEval & Davidson & Founta  \\ 
\hline
AbuseEval             & \textbf{0.634}     & 0.466    & 0.582   \\
Davidson              & 0.477     & \textbf{0.752}    & 0.633   \\
Founta                & 0.544     & \underline{0.582}    & \textbf{0.738}   \\ 
\hline
Two out-of-domain datasets & \underline{0.571} & 0.577 & \underline{0.640} \\
\hline
\end{tabular}
\caption{BERT performance in macro-averaged F1 scores. In-dataset training and evaluation is \textbf{bolded}; the best cross-dataset performance is \underline{underlined}. ``Two out-of-domain datasets'': When the training sets of two datasets apart from the evaluation dataset are combined for training. }
\label{tab:macro_f1}
\end{table}

\subsection{The impact of keywords on in-dataset classification}

\begin{figure}
    \centering
    \includegraphics[width=\columnwidth]{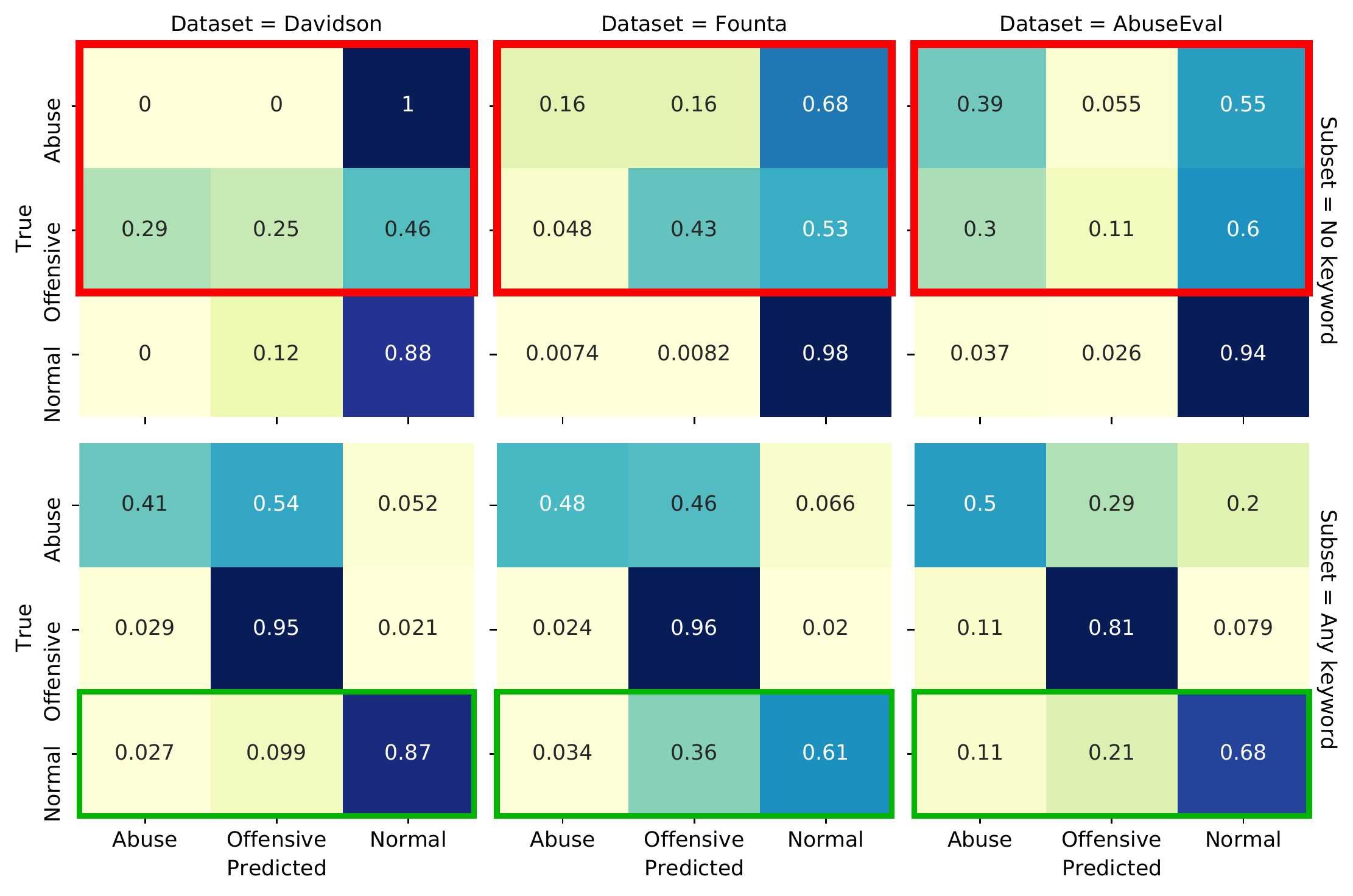}
    \caption{Confusion matrices of BERT predicted vs. true class labels, normalised by the true values, organised by keyword presence and datasets. True abusive and offensive instances in the ``no keyword'' subset are implicit expressions (red boxes); True normal instances in the ``any keyword'' subset are non-abusive use of keywords (green boxes). Models are trained and evaluated on the same datasets. }
    \label{fig:confusion}
\end{figure}

We first discuss results on in-dataset settings, i.e. were different subsets of the same dataset are used for training and testing. Figure \ref{fig:confusion} shows, for each class label, how having or not having keywords impacts what the classification model predicts. The red boxes highlight cases of implicit abuse and offensive language, whereas the green boxes highlight cases with non-abusive use of keywords. The three datasets have a lot in common, when it comes to what mistakes the model would make.  Moreover, these common patterns have a strong connection to the dependency between keyword presence and class labels shown in Table \ref{tab:presence}. We discuss these results next, first focusing on implicit and explicit expressions, and then on non-abusive posts with keywords present.

\subsubsection{Implicit vs explicit expressions}

On all three datasets, without keywords, the strong tendency to predict anything as harmless does not differ much across the three datasets, although, the datasets vary by how frequently instances contain keywords  (intensity) and how strong the association between having keywords and the abuse and offensive labels is (dependency). It is likely that the effect of these two factors offset each other -- recall that \textit{Davidson} is the least keyword-dependent, which is expected to be an advantage, but it is also the most keyword-intensive, limiting the available implicit instances to learn from.

At the same time, the relative performance on the implicit abuse and offensive classes depends on the imbalanced class ratios in the datasets. For both \textit{Davidson} and \textit{Founta}, where there are many more offensive than abuse instances, the model struggled more with implicit abuse than implicit offensive language. In contrast, in \textit{AbuseEval}, where there are many more abuse than offensive instances, the proportion of correctly classified implicit abuse is much larger. 

In clear contrast to the implicit case, explicit abuse and offensive language are much easier to detect. Nonetheless, the model tends to misclassify explicit abuse as offensive language. This problem is shared by all datasets regardless of the proportion of abuse. 
It is linked to the proportion of explicit instances comparing the abuse and offensive classes: in all three datasets, offensive language contains keywords much more frequently than abuse. 
The severity of the problem depends on how big this frequency contrast is. On \textit{Davidson}, offensive language is almost 10 times more likely to be explicit than abuse, the model made this mistake most frequently, followed by \textit{Founta} and then \textit{AbuseEval}.

\subsubsection{Non-abusive use of keywords}

On instances with non-abusive use of keywords (intersection of Any keyword \& Normal), as expected, the model displays a tendency to falsely flag innocent normal speech as offensive or abuse, although this tendency is not as strong as the effect of implicitness on identifying abuse.

Generally, the model tends to mistake these misleading instances as offensive rather than abuse. Offensive language being explicit much more frequently than abuse underlies this phenomenon.

How keyword-dependent the dataset is has a clear effect on how common the model falsely flags non-abusive keyword use as offensive or abuse. On the least keyword-dependent \textit{Davidson}, where there are a lot of instances with non-abusive keyword use, such mistakes are much rarer than in the other two datasets, and the recall of the normal class is similar with or without keywords. The model makes such mistakes more frequently on the most keyword-dependent \textit{Founta}, followed by \textit{AbuseEval}, even though both datasets have smaller offensive ratios than \textit{Davidson}.

In summary, the in-dataset performances reveal that the dependency between class labels and keyword presence is the biggest factor underlying the main challenges in classification models trained and evaluated end-to-end. 

\subsection{Factors in cross-dataset generalisation}

\begin{figure}[htp]
    \centering
    \subfigure[Single training dataset]{\includegraphics[scale=0.24]{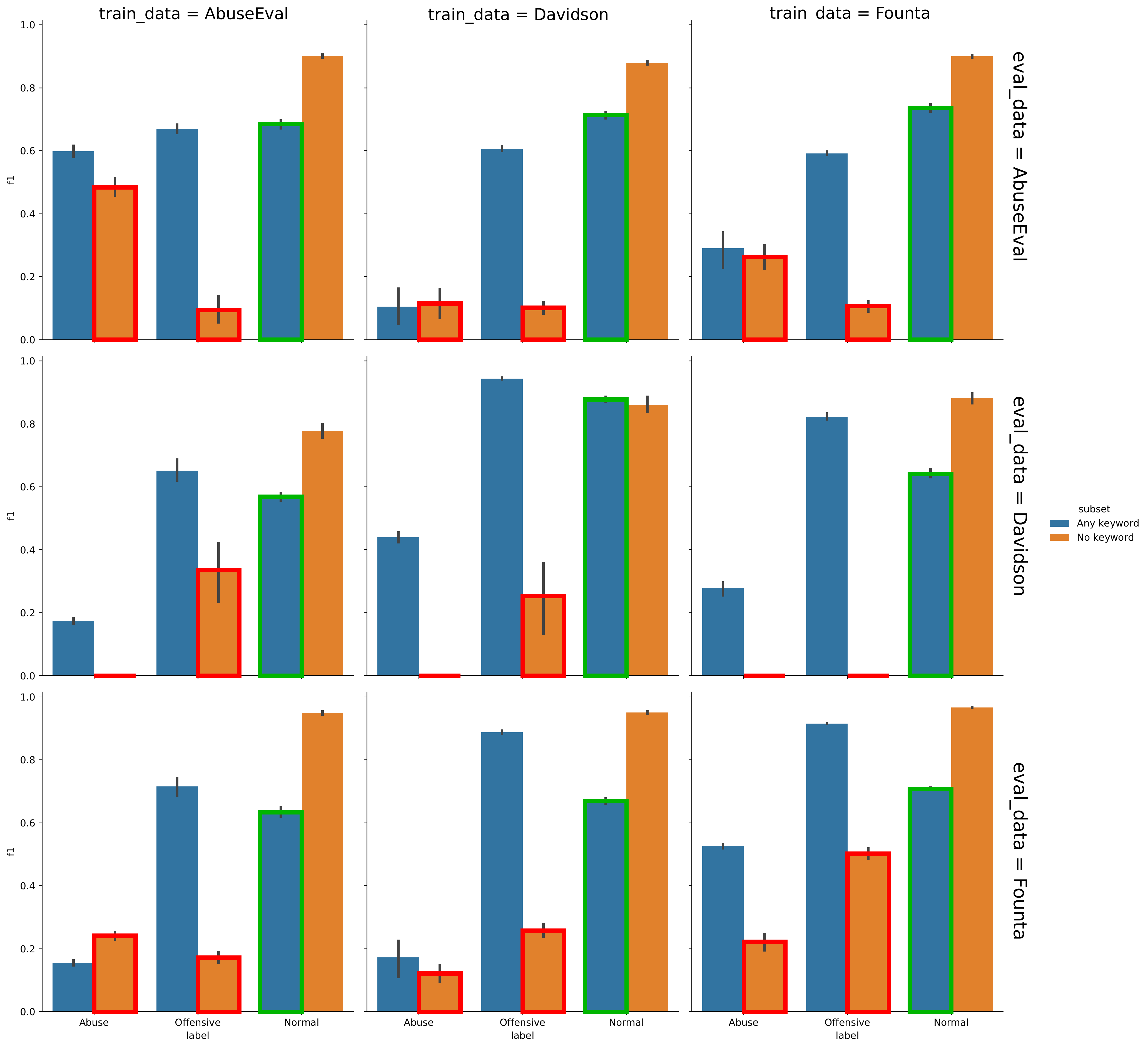}\label{subfig:1v1}}
    \subfigure[Double training dataset]{\includegraphics[scale=0.241]{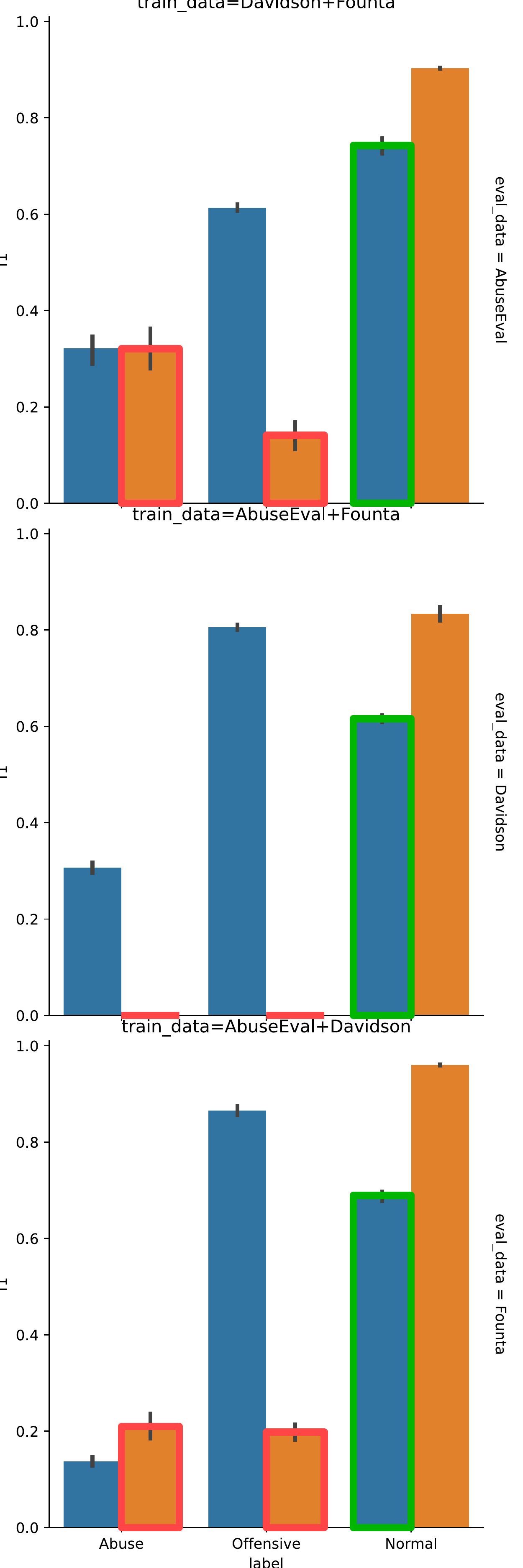}\label{subfig:2v1}}
    \caption{BERT model's performance, in- and cross-dataset evaluation on three datasets, on instances with or without keywords, by class labels. Showing 95\% confidence intervals generated through 1000 iterations of bootstrapping. Bars surrounded by a red box refer to cases of implicit abuse and implicit offensive language, whereas those surrounded by a green box refer to non-abusive use of keywords.}
    \label{fig:base_cross_f1s}
\end{figure}

Figure \ref{fig:base_cross_f1s} breaks down how the model generalises to a different dataset on each class label with or without keywords, compared to in-dataset evaluation (top left, middle, bottom right). Red boxes highlight performances on implicit abuse and implicit offensive language, and green boxes highlight performances on normal, non-abusive posts with keywords.

\subsubsection{Generalisation difficulty vs detection difficulty}

In the scenario of cross-dataset evaluation, the model also struggles with implicit expressions and non-abusive keyword use, although the extents differ. When the model is evaluated on an unseen dataset, it is additionally faced with the difference in class label definitions, manifested in the gap between in- and cross- performances.

Consistently, normal speech without any keywords is the easiest for the model to generalise on across datasets. 
Generalisation on explicit offensive language is noticeably more difficult than in-dataset evaluation.
When it comes to implicit expressions, offensive language or abuse, the model struggles to both detect and generalise. Abuse carries additional challenges: the difficulty to detect and generalise posed by implicitness is even more severe than the offensive case; even the explicit cases are hard to detect and generalise on. 

\subsubsection{Generalisation and similarity factors between datasets}

In \S \ref{subsubsec:implication_training}, we discussed how sampling and annotation approaches caused datasets to have varying degrees of keyword-intensity and keyword-dependency, as well as varying specificity of ``abuse'' definitions. Table \ref{tab:characteristics} summarises how the three datasets compare on these factors. 
As hypothesised, generalisation depends on dataset similarity on these factors, reflected in macro-averaged F1 scores shown in Table \ref{tab:macro_f1}.

Breaking down generalisation by the types of instances (Figure \ref{fig:base_cross_f1s}), we see that, while keywords affect the relative performance on implicit and explicit expressions, the coverage of the ``abuse'' class limits performance on this class. 

\begin{table}[ht]
\small
\centering
\begin{tabular}{p{1.7cm}|p{3cm}p{3cm}p{3cm}}
\hline
          & Keyword intensity & Keyword dependency & Abuse definition coverage  \\ 
\hline
AbuseEval & +                 & ++                 & ++++                        \\
Davidson  & ++++               & +                  & +                          \\
Founta    & ++                & ++++                & ++                         \\
\hline
\end{tabular}
\caption{Comparison of dataset characteristics contributing to generalisation. Summarised from \S \ref{subsubsec:implication_training}.}
\label{tab:characteristics}
\end{table}

Having more instances of a certain type of speech is related to better generalisation for that specific type for each dataset, such as explicit offensive language in \textit{Davidson}.
Similarity in definitions is linked to better generalisation between dataset pairs, as seen between \textit{Davidson} and \textit{Founta}. Nonetheless, the higher keyword-dependency in \textit{Founta} specifically degrades implicit offensive language.
There is a one-way generalisation advantage on implicit offensive language from less keyword-dependent data to higher ones -- on implicit offensive language, the \textit{Davidson}-trained model also outperformed the \textit{AbuseEval} one. 

\subsubsection{Generalisation from heterogeneous training data}

Previous research showed that heterogeneous augmentation (augmenting in-domain training data with out-of-domain training data) can be detrimental \citep{glavas_xhate-999_2020};  1-to-1 cross-dataset generalisation largely depends on similarity in keyword distribution and abuse class definition, reflected in implicit and abuse class performances. 
Here, we experiment with heterogeneous training data without in-domain data (combining two datasets that do not include the evaluation dataset) to connect these two observations.

Looking at the highest level of macro-averaged F1 scores (Table \ref{tab:macro_f1}), combining two sources of out-of-domain data mostly results in an improvement over generalisation from only one out-of-domain training dataset, except for when both training datasets generalise poorly individually -- on \textit{Davidson}, where the gaps between cross-dataset and in-dataset performance is the largest.

Subset performance (Figure \ref{subfig:2v1}) shows that the improvement is on the most difficult cases for generalisation -- more so on the abuse class than on implicit expressions. While an abuse class with a narrow definition generalises poorly to a broader one in one-to-one scenarios, when two narrow definitions are combined, the improvement is evident -- comparing one-to-one and two-to-one generalisation on \textit{AbuseEval} as the evaluation dataset. If one dataset is significantly larger (\textit{Founta}), relative subset performances would resemble the one-to-one generalisation from the larger dataset.

\subsection{Discussions: hypothesis verification and additional insights from model behaviour}

In \S \ref{ssec:discussion1}, we discussed how sampling and annotation approaches affect key characteristics of a dataset: keyword-intensity, keyword-dependency, and class proportions. We also hypothesised how each characteristic would manifest in detection and generalisation of instances of different types. Here, through the model behaviour expressed in in- and cross-dataset experiments, we complete the hypothesis verification and discuss additional insights: the relevant importance of the characteristics, and how they interact with each other. Through these analyses, we provide suggestions on dataset selection and construction, in order for better detection and generalisation of the misleading expressions -- implicit expressions and non-abusive keyword use.

\subsubsection{The relative effect of and interaction between keywords and class labels}

Firstly, we hypothesised that the models would have the biggest difficulty when detecting instances with implicit expressions of abuse or offensive language and non-abusive keyword use, as keyword use is associated with the offensive and abuse classes. 
Secondly, higher keyword-intensity and keyword-dependency were both expected to worsen these two challenges. 
We additionally hypothesised that the models will perform the best on the respective majority classes of the training datasets, as a general pattern of machine learning models.

Indeed, models show common struggles with the two challenging types of instances, confirming our first hypothesis: the models tend to mistake implicit abuse and offensive language as normal speech, and normal speech with non-abusive keyword use as offensive or abusive. However, the effects are not equal. The models are much more likely to miss the implicit expressions in abuse or offensive language than to falsely flag normal speech with keywords as offensive or abuse. This means that, in detecting abuse and offensive language, the absence of strong indicative lexical features has a stronger effect on causing false negatives than the presence of them on causing false positives.

As expected, being highly keyword-dependent is always detrimental. The model learns a stronger association between keywords and offensive classes, hindering the classification of both implicit expressions and non-abusive keyword use, as shown in the model behaviour when training and evaluated on \textit{Founta}. 

By contrast, being highly keyword-intensive can be a double-edged sword, under the influence of keyword-dependency. On one hand, it limits the total instances without keywords available for training. As a result, the model performs poorly on the implicit expressions of abuse and offensive language, as shown in the results for \textit{Davidson}, despite it being the least keyword-dependent dataset. Furthermore, the model can even mislabel normal speech without keywords as offensive, which the other two models seldom do on the other two datasets. On the other hand, it benefits the detection of non-abusive keyword use and explicit offensive language. If and only if, with suitable annotation instructions, the dataset's keyword-dependency is low, containing more keywords can mean having more instances with non-abusive keyword use, which facilitates the classification of such instances. The \textit{Davidson} model displays such a clear advantage over the other two, which do not differ a lot on this aspect. 
In binary classification, explicitness benefited the detection of the positive class \citep{graumas_twitter-based_2019}. Here, however, having more explicit expressions only benefits the detection of explicit offensive language -- but not explicit abuse.

The reason why explicitness does not benefit the detection of abuse in the same way as offensive language lies in the fact that the latter is always more likely to be explicit. As a result, all three models struggle with the ``abuse'' class the most, and in similar ways. When there are keywords, explicit abuse is often mistaken with explicit offensive language, even on \textit{AbuseEval}, where there is more abuse than offensive language. When there are no keywords, implicit abuse is most commonly mistaken as harmless, normal speech, as is implicit offensive language.

The effect of class labels is mainly through the coverage of the abuse class, rather than majority vs minority. This coverage is reflected in the dataset as the relative proportions between abuse and offensive language. 
The lower the ratio of abuse, the more likely the model is to mistake explicit abuse as offensive language. This is seen across the three datasets. Furthermore, on the extremely challenging instances with implicit expressions, the smaller class is misclassified even more: on \textit{Founta} and \textit{Davidson}, implicit abuse was misclassified relatively more often than implicit offensive language, while it was the opposite for \textit{AbuseEval}.
This offers an explanation to the inconsistency in the literature on the effect of the positive class ratio in binary classification: the proportion of actual abuse is likely a moderating factor.

\subsubsection{Similarities and differences between generalisation and in-domain detection}

In terms of model generalisability, our hypothesis was mainly based on the overall similarity between datasets: generalisation between the two most different datasets overall would be the most difficult. This similarity consisted of three key dataset characteristics that are direct products of sampling and annotation approaches: keyword intensity, keyword dependency, and the abuse class coverage. 

This hypothesis is verified through cross-dataset macro-averaged F1 scores, but a breakdown of the performance by keywords and classes produced far more insights, on the connection between detection and generalisation and on the separate effects and relative importance of the three characteristics.

Instances that are harder to detect are also mostly harder for the model to generalise on, but there are exceptions. The difficulty of detecting implicit expressions and abuse seen in the in-dataset scenario is magnified when it comes to cross-dataset generalisation. However, generalisation on explicit offensive language, which is a lot easier to detect, is similarly difficult to that on instances with non-abusive keyword use. This means that while all datasets show an unequivocal, strong association between keywords and the offensive class, this association is slightly different across datasets. On the other hand, while non-abusive use of keywords misleads all models, the way in which it does so has limited variability.

As in the in-dataset case, keyword-dependency is always detrimental for generalisation, which mainly affects the implicit expressions. When keyword intensity is also low, as in \textit{AbuseEval}, it means there is more training instances for implicit expressions, facilitating generalisation on such instances. Even in situations where both training and evaluation datasets are highly keyword-intensive, as in \textit{Founta} and \textit{Davidson}, the less keyword-dependent one, \textit{Davidson}-trained model, generalises better to \textit{Founta} on the implicit instances than the other way round.

The benefit of keyword-intensity for generalisation on non-abusive keyword use and explicit offensive language is similar to that in the in-dataset scenario. Specifically, generalisation on these instances is the best between the more keyword-intensive \textit{Davidson} and \textit{Founta}, which were sampled in similar ways. On these instances, they also generalise better to \textit{AbuseEval} than the other way round. This benefit of a wider coverage of keywords outweighs the similarity in keyword topic distributions.
Previous research found that wider coverage of phenomena improves generalisability \citep{pamungkas_cross-domain_2019, nejadgholi_cross-dataset_2020}; this finding adds keywords to the type of such phenomena whose coverage can benefit generalisability -- limited to instances that actually contain keywords.

This comes at a cost on the implicit expressions. Depending on the proportion of such instances, this demerit can be hidden by the smoke screen of a good performance on the explicit ones, when looking at the overall macro-averaged F1 scores. Because \textit{Davidson} and \textit{Founta} are both heavy in explicit instances, the overall F1 scores are dominated by the fact that they generalise better to each other on the explicit instances, while the advantage on the implicit ones displayed by \textit{AbuseEval} is not reflected in the overall F1 scores. 
Only looking at the macro-averaged F1 is also the reason why previous research concluded that having more explicit expressions carries generalisability benefit \citep{wiegand_detection_2019}, while in reality, such benefit is limited to the dominating explicit instances.

The abuse class again limits the above effect, similar to the in-dataset scenario. Previous work in binary classification suggested that a positive class with a broad definition is more generalisable \citep{nejadgholi_cross-dataset_2020, fortuna_how_2021}. This is mostly true in our results, but when the narrow ones are sufficiently similar, they generalise reasonably well to each other.

The above effects translate to the scenario when heterogeneous out-of-domain data is combined for training, with the additional benefit of better generalisation compared to single source. Having multiple data sources likely served as regularisation against overfitting to one dataset. This benefit is nonetheless limited by the pair-wise dataset similarity and one-to-one generalisation, and is specific for cross-dataset generalisation, as augmenting in-domain training data with out-of-domain training data can be detrimental \citep{glavas_xhate-999_2020}. 
Combining multiple sources helps alleviate the difficulty of generalising from narrow to broader definitions of abuse.

\subsubsection{Suggestions on dataset construction and application} 
In \S \ref{ssec:discussion1}, we discussed how annotation and sampling contribute to dataset characteristics. In this section so far, we saw that these characteristics of a dataset in turn affect model performance on different types of instances.

Drawing from these discussions, attention should be paid on both sampling and annotation when constructing a dataset, in order for better, generalisable model performance on the most challenging instances, i.e. implicit expressions and non-abusive keyword use.

To start with, the initial sample is better to be not drawn with biased sampling directly using slurs and profanity. Otherwise, such a filtering criterion increases the frequency of slurs and profanity, which in turn impairs model performance on the implicit expressions. Furthermore, it reduces the instances actually containing abuse in the sample, so this is especially important when studying abuse as a separate phenomena from offensive language.

If words and phrases are to be used to boost the ratio of the abusive class, using those that are not inherently offensive or abusive  \citep{zampieri_predicting_2019} can reduce the direct link between keywords and class labels. Furthermore,  boosted random sampling (applying the criteria after drawing an initial random sample) results in less biased data than biased sampling (drawing a biased sample with the criteria) \citep{wiegand_detection_2019}. 
Instead of keyword-based biased sampling, There are a few alternatives which can reduce the bias towards slurs and profanities. 
One possible approach is to draw samples based on communities, such as forums which are banned due to hateful discussions \citep{vidgen2021introducing, de_gibert_hate_2018}. Another alternative is to use semi-supervised learning, such as the SOLID \citep{rosenthal2021solid} dataset, labelled with confidence scores of models trained on \textit{OLID} \citep{zampieri_predicting_2019}. By using a range of models with different inductive biases and the means and standard deviations of the confidence scores, the semi-supervised labels are not overfitted to any particular model. These methods that do not rely on keyword filters removes one significant source of potential bias. Nonetheless, these models are not free from additional risks: some narrow communities may have very specific language styles which deviate from the mainstream or require substantial contextual knowledge in order to understand certain discussions; biases in the seed dataset may be transferred to the semi-supervised data.
These approaches can also be combined to reduce the influence of a small number of major biases. 

Yet, it is worth keeping in mind that the challenge brought by keywords also comes from the nature of abusive language, rather than the sampling method -- offensive language is always more likely to contain keywords than abuse. Thus, no matter how sophisticated the sampling process is, the ratio of implicit and explicit instances and author and topic distributions should be carefully studied, controlled, and reported.

Annotation guidelines should be as specific as possible. A prerequisite is to have detailed definitions. Confusions between perceived offensiveness and the specific abuse under investigation can be reduced by explicitly listing out common mistakes in the guidelines \citep{basile_semeval-2019_2019}. Biases against language styles should be controlled by explicitly asking annotators to rely on words less \citep{davidson_automated_2017} and reminding them of factors in dialects \citep{sap_risk_2019}. If high levels of details appear overwhelming, a decision-tree-like guideline \citep{caselli_i_2020} can simplify the decision process. 

The above applies to general dataset construction. Additionally, depending on the specific application scenario, there are other factors to consider.

The target domain can be rich in slurs and/or profanity, such as a community of marginalised groups that frequently use reclaimed slurs. In these communities, false positives caused by slurs carry serious negative social impact -- further oppressing marginalised groups. Thus, data for training an abusive language detection model need a balance between good coverage of such keyword use and implicit expressions. In such a scenario, sampling directly through slurs and/or profanity can be actually beneficial,  provided that extra care is taken to make sure there are enough implicit instances.

The definition of abuse should also be tailored to knowledge about the target domain. Having a wide coverage of all kinds of abuse in the training data is not as useful as having an accurate representation of what abuse looks like in the evaluation data. Thus, if there is enough knowledge about the abuse present in the target domain, the definition when building or choosing the training data should be as specific as possible, such as ``hate speech''. Otherwise, when there is much unknown about the target domain, using broadly defined abuse for training is a safe choice.

\section{Conclusions}

By looking at the presence of keywords categorised as profanity or slurs in datasets, in this paper we study the tendency of abusive language detection models to label content as abusive, offensive or normal. Investigating with three widely-used abusive language datasets where this 3-way classification is possible, we assess the implications of decisions made in the dataset construction stage in the development of abusive language detection models. We break our analysis into two main parts. First, we analyse the prominence of profanity and slurs in the different datasets, focusing both on keyword intensity (presence of keywords in the dataset) and keyword dependency (association of keywords with classes). And second, we assess the impact of these dataset patterns on the ability of models to detect abusive language, both in in- and cross-dataset scenarios, looking at the ability to generalise across datasets. We focus on two challenging cases in detail: (1) implicit cases of abusive and offensive language where keywords aren't present, and (2) non-abusive use of keywords.

We defined three research questions for our study, which we answer next. 

\textbf{RQ1: How do approaches in dataset construction lead to different patterns of slurs and profanity presence in the dataset? }

In sampling, factors that make the process more biased towards keywords are: biased rather than boosted random sampling, text search through slurs and profanity rather than terms that are not inherently offensive. A sampling process more biased towards keywords makes a dataset contain more keywords, i.e. more keyword-intensive. Less biased processes lead to similar topic distributions of the keywords.

In annotation, factors that make the process more biased towards keywords are: crowd-sourced rather than expert annotators, brief rather than detailed instructions. An annotation process more biased towards keywords makes the association between keywords and offensive and abuse class labels stronger, i.e. more keyword-dependent. There are also common patterns in how abusive keywords of different topics are considered.

Additionally, both sampling and annotation affect class distributions. Sampling biased towards keywords increase offensive instances but decreases the ratio of abuse compared to offensive. A broader definition of abuse in annotation leads to wider coverage of phenomena. 

\textbf{RQ2: How does such keyword presence in turn affect the detection model behaviour?}

The role of keywords is two-folds: their association with class labels make them act as useful lexical features, which also means that instances that go against this association lack training instances.
Thus, the most challenging instances for classification are the ones with fewer lexical features and instances available in the training data: implicit expressions and abuse of any kind, followed by non-abusive keyword use.

Comparing datasets, varying keyword intensity and dependency affects performance through affecting the number of instances. Keyword dependency impairs the detection of implicit expressions and non-abusive keyword use. Keyword intensity decreases performance on implicit expressions, but improves that on explicit ones. Provided with low keyword dependency, it can also benefit the classification of non-abusive keyword use.

All models make similar mistakes when it comes to abuse, mistaking it either as offensive or normal speech depending on whether keywords are present, although the coverage of the abuse class in the training data has some influence.

Depending on the makeup of the evaluation dataset, some of these effects may not be reflected in the overall model performance.

\textbf{RQ3: How does this effect differ when it comes to model generalisation?}

The above effects of keywords apply to both in-dataset detection and cross-dataset generalisation. 

Generalisation introduces extra challenge: the effects of keywords are more prominent, especially on implicit expressions and abuse; generalisation on the abuse class largely depends on the definition difference between datasets. 

Combining out-of-domain data is likely to be beneficial for generalisation compared to training on single-source out-of-domain data, mainly by addressing the challenge of abuse definition differences.

Based on the above research questions and answers, we have also provided suggestions on dataset construction.

~\\

\textbf{Limitations and future work.} Our work is not free from limitations, which also open up new directions for future research. 
We had kept the experiment conditions consistent across datasets and taken a careful hypothesis verification approach to explain the results. Our findings were also in line with available relevant research. Nonetheless, even though more rigorous hyperparameter tuning could potentially lead to marginally better performance metrics, there are random peripheral factors which would have fluctuated the results.
The arguably most important limitation is the definition of implicit and explicit based on slurs and profanity. We used lexica to define explicitness, for which the motivation was two-fold: (1) it operationalises explicitness without inconsistency and subjectivity which would have otherwise been introduced by manual annotation, by biases in machine learning models or by semi-supervised approaches such as using the Perspective API; (2) our use of lexica is consistent with previous literature, putting our findings in context \citep{waseem_understanding_2017,caselli_i_2020,wiegand-etal-2021-implicitly-abusive}. However, implicitness and explicitness are highly subjective notions \citep{vidgen_detecting_2020}. Thus, the operational definition we use may not match the implicitness and explicitness perceived by humans in ambiguous cases, such as the use of slurs that also have a non-hateful meaning, or expressions through violent words such as ``hate'' or ``kill''. There are also instances where humans would not agree on the nature of the speech, such as ``... are dumb''. Thus, in reality, gold standards of what constitutes explicit and implicit hate are not always possible. It is important that the readers bear in mind that our definition of explicitness is a widely used approximation of something whose ground truth does not exist, and focus on the qualitative differences in the performance metrics rather than taking them at face value.

We compared three real, widely used multi-class datasets, which enabled our findings to be applicable to studies that have used these datasets, but also extensible to other scenarios with datasets with similar characteristics. From the perspective of experimental study, confounders, such as the size of data, can be better controlled through subsampling \citep{razo_investigating_2020}. Nonetheless, we expect data size to have had limited influence on our findings, as previous research in abusive language detection showed that its effect is limited compared to other factors \citep{nejadgholi_cross-dataset_2020, fortuna_how_2021}.

Because of our focus on a more fine-grained three-way classification task as opposed to the widely used binary formulation of abusive language detection, and keeping a consistent definition of implicitness across datasets to enable cross-dataset experiments, we could not make full use of some very relevant human annotations. We encourage future research to empirically study human-annotated implicitness in the original \textit{AbuseEval} \citep{caselli_i_2020} and datasets on distinguishing abusive and non-abusive use of swearing \citep{pamungkas_you_2020} and implied statements of implicit hate \citep{elsherief-etal-2021-latent}. Some binary datasets also have hierarchical sub-categories, such as targeted or untargeted \citep{zampieri_semeval-2019_2019,mandl_overview_2019}, group-directed or individual directed \citep{ zampieri_semeval-2019_2019, fersini_overview_2018}. In principle, a similar cross-dataset empirical analysis can be applied to these hierarchical labels where categories match. When sub-categories do not perfectly align across datasets, cross-dataset experiments can be enabled by formulating overlapping sub-categories as a binary task, such as ``Sexual Harassment \& Threats of Violence'' in \cite{fersini_overview_2018} and ``Sexual Violence'' in \cite{rodriguez2021overview}. 

Finally, our findings highlight the challenge of implicit expressions --compared to explicit ones--, and abuse --compared to offensive language--, which are not always reflected in the overall performance metrics. However, these are the instances that carry the most practical implications. We thus provide suggestions on sampling and annotation for future dataset construction.
Additionally, future research that build abusive language detection models to optimise, carefully investigate model performance, or motivate model designs considering also the most challenging instances.

\bibliographystyle{elsarticle-num} 
\bibliography{main,fixed_hate_speech_synced}

\end{document}